\documentclass{article}

\PassOptionsToPackage{numbers}{natbib}

\usepackage[preprint]{nips_2018}

\usepackage[utf8]{inputenc} 
\usepackage[T1]{fontenc}    
\usepackage{hyperref}       
\usepackage{url}            
\usepackage{booktabs}       
\usepackage{amsfonts}       
\usepackage{nicefrac}       
\usepackage{microtype}      
\usepackage{dsfont}
\usepackage{verbatimbox}
\usepackage{listings}
\usepackage{capt-of}
\usepackage{graphicx} 
\usepackage{fancyvrb}

\usepackage{caption}
\usepackage{subcaption}

\fvset{fontsize=\footnotesize}

\usepackage{multirow}
\usepackage{array}

\usepackage{amssymb}
\usepackage{listings}
\usepackage{wrapfig}
\usepackage{tabularx}

\usepackage{verbatim}
 \usepackage{booktabs}
\usepackage{algorithm}
\usepackage{algorithmic}
\usepackage{tikz}
\usetikzlibrary{fit}
\usetikzlibrary{arrows.meta}
\usetikzlibrary{positioning}
\usetikzlibrary{decorations.text,decorations.pathreplacing}
\usetikzlibrary{decorations.pathmorphing}

\usepackage{amsmath}
\usepackage{hyperref}
\DeclareMathOperator*{\argmax}{arg\,max} 

\newcommand{\expect}{\mathds{E}} 
\newcommand{\probability}{\mathds{P}} 
\newcommand{\indicator}{\mathds{1}} 

\newcommand{\exampleImageSize}{2cm}

\usepackage{amsthm}
 
\theoremstyle{definition}

\title{Learning to Infer Graphics Programs from Hand-Drawn Images}

\author{Kevin Ellis\\MIT\\\texttt{ellisk@mit.edu}\\
 \And
 Daniel Ritchie\\Brown University\\\texttt{daniel\_ritchie@brown.edu}\\
 \AND
 Armando Solar-Lezama\\MIT\\\texttt{asolar@csail.mit.edu}\\
 \And
 Joshua B. Tenenbaum \\MIT\\\texttt{jbt@mit.edu}\\
}

\begin{document}

\maketitle

\begin{abstract}
  We introduce a model that learns to convert simple hand drawings
  into graphics programs written in a subset of \LaTeX.~The model
  combines techniques from deep learning and program synthesis.  We
  learn a convolutional neural network that proposes plausible drawing
  primitives that explain an image. These drawing primitives are a
  specification (spec) of what the graphics program needs to draw.  We
  learn a model that uses program synthesis techniques to recover a
  graphics program from that spec. These programs have constructs like
  variable bindings, iterative loops, or simple kinds of
  conditionals. With a graphics program in hand, we can correct errors
  made by the deep network, measure similarity between drawings by use of similar high-level geometric structures, and extrapolate drawings.
\end{abstract}
\vspace{-0.3cm}
\section{Introduction}

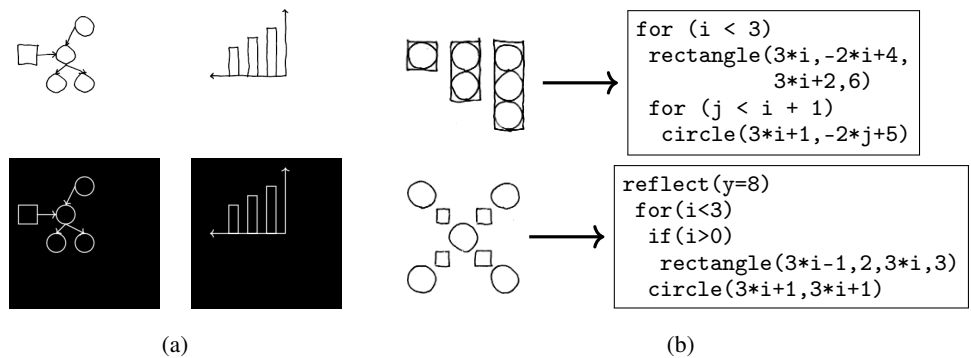
\begin{figure}[H]
  \begin{minipage}[b]{0.35\linewidth}  \centering
\begin{tabular}{ll}
  \includegraphics[width = \exampleImageSize]{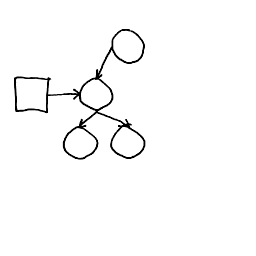}&
    \includegraphics[width = \exampleImageSize]{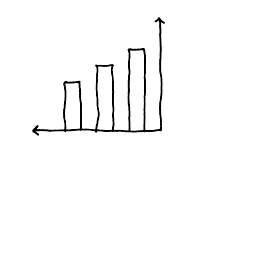}\\
  \includegraphics[width = \exampleImageSize]{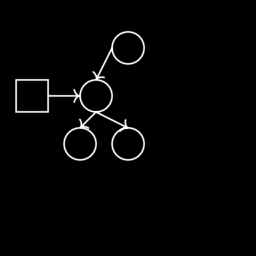}&
    \includegraphics[width = \exampleImageSize]{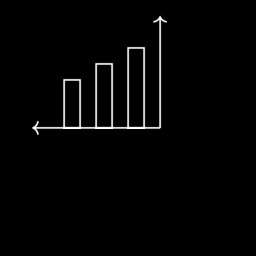}
\end{tabular}
\subcaption{}
  \end{minipage}
  \begin{minipage}[b]{0.6\linewidth}\centering
    \begin{tikzpicture}
      \node(picture1) at (0,-1) {\includegraphics[height = 1.5cm]{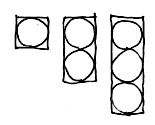}};
      \node(picture2) [below = 0.3cm of picture1] {\includegraphics[height = 1.5cm]{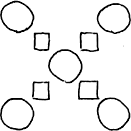}};
      \node[right=1cm of picture1] (c1) {\fbox{\firstFirstPageCode}};
      \node[right=1cm of picture2] (c2) {\fbox{\secondFirstPageCode}};
      \draw[very thick,->] (picture1.east)  -- (c1.west);
      \draw[very thick,->] (picture2.east)  -- (c2.west);
    \end{tikzpicture}
    \subcaption{}
  \end{minipage}
  \caption{(a): Model learns to convert hand drawings (top) into \LaTeX~(rendered below). (b) Learns to synthesize high-level graphics program from hand drawing.}\label{firstPageExamples}
\end{figure}

Human vision is rich -- we infer shape, objects, parts of objects,
and relations between objects -- and vision is also abstract:
we can perceive the radial symmetry of a spiral staircase,
the iterated repetition in the Ising model,
see the forest for the trees, and also the recursion within the trees.
How could we build an agent with similar visual inference abilities?
As a small step in this direction, 
we cast this problem as program learning,
and take as our goal to learn high--level
graphics programs from simple 2D drawings.
The graphics programs we consider make figures like those found in machine learning papers
(Fig.~\ref{firstPageExamples}),
and capture high-level features like 
symmetry, repetition, and reuse of structure.


\begin{myverbbox}[\small]{\firstFirstPageCode}
for (i < 3)
 rectangle(3*i,-2*i+4,
           3*i+2,6)
 for (j < i + 1)
  circle(3*i+1,-2*j+5)
\end{myverbbox}
\begin{myverbbox}[\small]{\secondFirstPageCode}
reflect(y=8)
 for(i<3)
  if(i>0)
   rectangle(3*i-1,2,3*i,3)
  circle(3*i+1,3*i+1)
\end{myverbbox}

 The key observation behind our work is that going from pixels to programs involves two distinct steps, each requiring different technical approaches. The first step involves inferring what objects make up an image -- for diagrams, these are things like as rectangles, lines and arrows. The second step involves identifying the higher-level visual concepts that describe how the objects were drawn. In Fig. 1(b), it means identifying a pattern in how the circles and rectangles are being drawn that is best described with two nested loops, and which can easily be extrapolated to a bigger diagram.

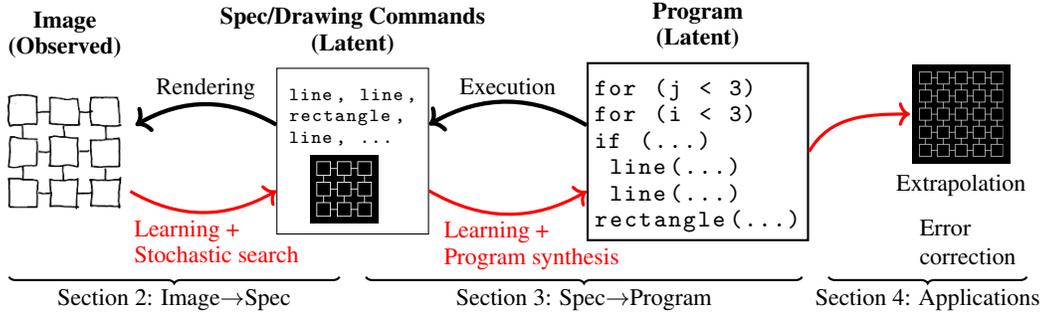
\begin{figure}[h]
  \centering\begin{tikzpicture}
  \node[ thick,anchor = west,inner sep=0pt,label={[yshift = 0.3cm]{\small \begin{tabular}{c}
          \textbf{Image}\\
          \textbf{(Observed)}
  \end{tabular}}}](observation) at (0,0) {\includegraphics[width = 1.5cm]{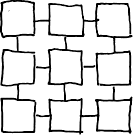}};
    \node[ultra thick,anchor = west,inner sep=0pt](traceSource) at (3.7,0.5){    \begin{lstlisting}[basicstyle = \scriptsize\ttfamily]
line, line,
rectangle,
line, ...
\end{lstlisting}};
    \node[ultra thick,anchor = west,inner sep=0pt](traceImage) at (4,-0.5) {
      \includegraphics[width = 0.9cm]{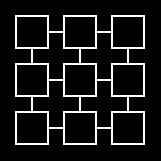}}; 
    \node(trace)[draw,thin,fit = (traceImage) (traceSource), label = above:{{\small \begin{tabular}{c}
            \textbf{Spec/Drawing Commands}\\
            \textbf{(Latent)}
    \end{tabular}}}] {};
    
    \node[draw, thick,anchor = west,inner sep=2pt,label=above:{\small \begin{tabular}{c}
          \textbf{Program}\\
          \textbf{(Latent)}
    \end{tabular}}](program) at (7.7,0) {
      \begin{lstlisting}
for (j < 3)
for (i < 3)
if (...)
 line(...)
 line(...)
rectangle(...)
    \end{lstlisting}};
    \node[ultra thick,anchor = west,inner sep=0pt,label=below:{\small Extrapolation}](extrapolate) at (12,0.5) {\includegraphics[width = 1.3cm]{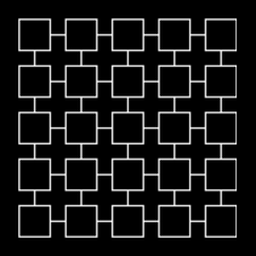}};
    \node[ultra thick,anchor = west,text width = 1.3cm,inner sep=0pt](errors) at (12.1,-1.2) {\small Error correction};

    \draw[->, ultra thick] ([yshift=10]trace.west) to[out = 150,in = 30] node[midway,yshift = 6]{{\small Rendering}} ([xshift=5,yshift=10]observation.east); 
    \draw[->, very thick, red] ([yshift = -15,xshift=3]observation.east) to[out = -30,in = -150] node[midway,yshift = -12,xshift=5]{{\small \begin{tabular}{l}
          Learning + \\Stochastic search
    \end{tabular}}} ([yshift = -15]trace.west);
    \draw[->, ultra thick] ([yshift=10]program.west) to[out = 150,in = 30] node[midway,yshift = 6] {{\small Execution}} ([yshift=10]trace.east);
    \draw[->, very thick, red] ([yshift = -15,xshift=0]trace.east) to[out = -30,in = -150] node[midway,yshift = -12,xshift=9]{{\small \begin{tabular}{l}
          Learning + \\Program synthesis
    \end{tabular}}} ([yshift = -15]program.west);

    \draw[->, thick, red, very thick] (program.east) to[out = 60,in = 180] (extrapolate.west);

    \draw[decoration = {brace,mirror,raise = 5pt},decorate,thick]
    ([yshift = -11,xshift = -130]trace.south) -- node[below = 6pt] {{\small Section~\ref{neuralNetworkSection}: Image$\to$Spec}} ([yshift = -11,xshift = -5]trace.south);
    \draw[decoration = {brace,mirror,raise = 5pt},decorate,thick]
    ([xshift = 5,yshift = -11]trace.south) -- node[below = 6pt] {{\small Section~\ref{programSynthesisSection}: Spec$\to$Program}} ([xshift = 170,yshift = -11]trace.south);
    \draw[decoration = {brace,mirror,raise = 5pt},decorate,thick]
([xshift = 180,yshift = -11]trace.south) -- node[below = 6pt] {{\small Section~\ref{applicationsSection}: Applications}} ([xshift = 255,yshift = -11]trace.south);
  \end{tikzpicture}
  \caption{Black arrows: Top--down generative model; Program$\to$Spec$\to$Image. {\color{red}Red} arrows: Bottom--up inference procedure. \textbf{Bold:} Random variables (image/spec/program)}\label{roadmap}
 \end{figure}

This two-step factoring can be framed as probabilistic inference in a generative model where a latent program
 is executed to produce a set of drawing commands,
 which are then rendered to form an image (Fig.~\ref{roadmap}).
 We refer to this set of drawing commands as a \textbf{specification (spec)} because it specifies what the graphics program drew while lacking the high-level structure determining how the program decided to draw it.
 We infer a spec from an image using stochastic search (Sequential Monte Carlo)
 and infer a program from a spec using  constraint-based program synthesis~\citep{solar2008program} --
 synthesizing structures like symmetries, loops, or conditionals.
 In practice, both stochastic search and program synthesis are
 prohibitively slow,
 and so we learn models that accelerate inference for both programs and specs,
 in the spirit of ``amortized inference''~\cite{paige2016inference},
 training a neural network to amortize the cost of inferring specs from images and using
 a variant of Bias--Optimal Search~\cite{schmidhuber2004optimal}
 to amortize the cost of synthesizing programs from specs.

 The new contributions of this work are (1) a working model that can infer high-level symbolic programs from perceptual input, and (2) a technique for using learning to amortize the cost of program synthesis, described in Section~\ref{learningASearchPolicy}.

\section{Neural architecture for inferring specs}\label{neuralNetworkSection}

We developed a deep network architecture for efficiently inferring a
spec, $S$, from a hand-drawn image, $I$.
Our model combines ideas from
Neurally-Guided Procedural Models~\citep{ritchie2016neurally}
and Attend-Infer-Repeat~\citep{eslami1603attend}, but
we wish to emphasize
that one could use
many different approaches from the computer vision toolkit to
parse an image in to primitive drawing commands  (in our terminology, a ``spec'')~\cite{nsd}.
Our network constructs the
spec one drawing command at a time, conditioned on what it has drawn so far (Fig.~\ref{architecture}).
We first
pass a $256\times 256$ target image and a rendering of the drawing commands so
far (encoded as a two-channel image) to a convolutional network. Given
the features extracted by the convnet, a multilayer perceptron then
predicts a distribution over the next drawing command to execute
(see Tbl.~\ref{drawingCommandTable}).
We also use a
differentiable attention mechanism (Spatial Transformer
Networks:~\cite{jaderberg2015spatial}) to let the model attend to
different regions of the image while predicting drawing commands.
We currently constrain
coordinates to lie on a discrete $16\times 16$ grid,
but the grid could be made arbitrarily fine. Appendix~\ref{architectureDetails} gives full architectural detail.  

\tikzset{>=latex}
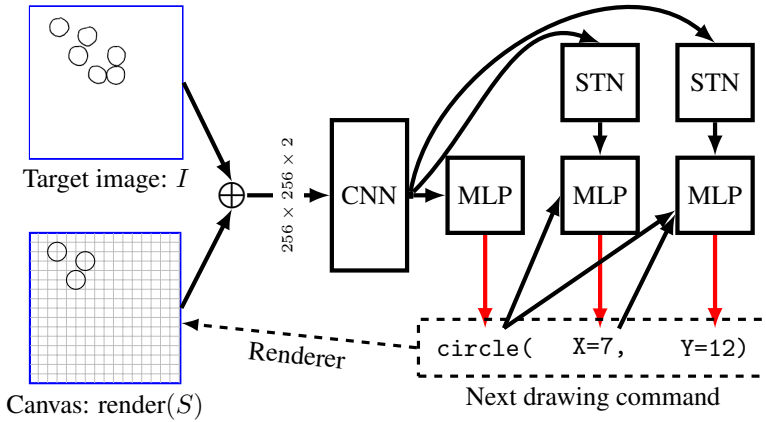
\begin{figure}[t]
  \begin{minipage}[c]{0.65\textwidth}
    \begin{tikzpicture}
  \node[draw,blue,ultra thick,anchor = west,inner sep=0pt,label=below:Target image: $I$](observation) at (0,-1) {\includegraphics[width = 2cm]{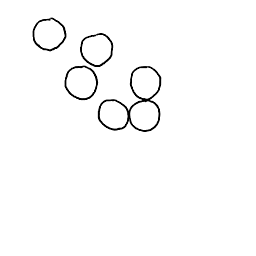}};
  \node[draw,blue,thick,anchor = west,inner sep=0pt,minimum width = 2cm,minimum height = 2cm,label=below:Canvas: render$(S)$] (canvas) at (0,-4) {};
  \draw[lightgray,ultra thin,step = 0.125] ([xshift = 0.5,yshift = 0.5]canvas.south west) grid ([xshift = -0.5,yshift = -0.5]canvas.north east);
  \draw (0.375cm,-3.25cm) circle (0.125cm);
  \draw (0.625cm,-3.625) circle (0.125cm);
  \draw (0.75cm,-3.375) circle (0.125cm);

  \node[draw,ultra thick,anchor = west,inner sep=0pt,minimum width = 1cm,minimum height = 2cm] (CNN) at (4,-2.5) {CNN};
  \node[inner sep = 0pt](tensorProduct) at ([xshift = -1.3cm]CNN.west) {$\bigoplus$};

  \node[draw,ultra thick,inner sep=0pt] (features) at ([xshift = 0.01cm]CNN.east) {};

  \node[draw,ultra thick,minimum size = 1cm](c1) at ([xshift = 1cm]features) {MLP};
  \node(l1) at ([yshift = -1.5cm]c1.south) {\verb|circle(|};
  \node[draw,ultra thick,minimum size = 1cm](a2) at ([xshift = 1cm,yshift = 1.5cm]c1.east) {STN};
  \node[draw,ultra thick,minimum size = 1cm](c2) at ([xshift = 1cm]c1.east) {MLP};
  \draw[->,ultra thick] (a2.south) -- (c2.north);
  \draw[->,ultra thick,red] (c1.south) -- (l1.north);
  \node(l2) at ([yshift = -1.5cm]c2.south) {\verb|X=7,|};
  \draw[->,ultra thick,red] (c2.south) -- (l2.north);
  \node[draw,ultra thick,minimum size = 1cm](a3) at ([xshift = 1cm,yshift = 1.5cm]c2.east) {STN};
  \node[draw,ultra thick,minimum size = 1cm](c3) at ([xshift = 1cm]c2.east) {MLP};
  
  \draw[->,ultra thick] (a3.south) -- (c3.north);
  \node(l3) at ([yshift = -1.5cm]c3.south) {\verb|Y=12)|};
  \draw[->,ultra thick,red] (c3.south) -- (l3.north);

  \draw[->,ultra thick] (features) -- (c1.west);
  \draw[->,ultra thick] (features) to[out = 45,in = 160] (a2.north);
  \draw[->,ultra thick] (features) to[out = 80,in = 140] (a3.north);
  \draw[->,ultra thick] ([xshift = 0.25cm]l1.north) -- (c2.west);
  \draw[->,ultra thick] ([xshift = 0.25cm]l1.north) -- ([yshift = -0.2cm]c3.west);
  \draw[->,ultra thick] ([xshift = 0.25cm]l2.north) -- ([yshift = -0.25cm]c3.west);

  \node(next)[draw,very thick,fit = (l1) (l2) (l3), dashed, label = below:{Next drawing command}] {};

  \draw[-{>[scale = 1.5]},very thick,dashed] (next.west) -- ([yshift = -0.2cm]canvas.east) node [midway, below, sloped] (TextNode) {Renderer}; 
  
  \draw[->,ultra thick] (canvas.east) -- (tensorProduct.south);
  \draw[->,ultra thick] (observation.east) -- (tensorProduct.north);
  \draw[->,ultra thick] (tensorProduct.east)  -- node[fill = white,rotate = 90] {{\tiny $256\times 256\times 2$}}  (CNN.west);
  \draw[-,ultra thick] (CNN.east) -- 
  (features);
  
\end{tikzpicture}
  \end{minipage}  \hfill%
  \begin{minipage}[c]{0.2\textwidth}
    \caption{Neural architecture for inferring specs from images. \textcolor{blue}{Blue}: network inputs. Black: network operations. \textcolor{red}{Red}: draws from a multinomial. \texttt{Typewriter font}: network outputs. Renders on a $16\times 16$ grid, shown in \textcolor{gray}{gray}. STN: differentiable attention mechanism~\citep{jaderberg2015spatial}.}  \label{architecture}
    \end{minipage}
\end{figure}

\begin{table}[h]
  \caption{Primitive drawing commands currently supported by our model.}
\label{drawingCommandTable}
\begin{tabular}{ll}\toprule
  \begin{tabular}{l}
    \verb|circle|$(x,y)$
  \end{tabular}& \begin{tabular}{l}
    Circle at $(x,y)$
    \end{tabular}\\
  \begin{tabular}{l}
    \verb|rectangle|$(x_1,y_1,x_2,y_2)$
  \end{tabular}&\begin{tabular}{l}
    Rectangle with corners at $(x_1,y_1)$ \& $(x_2,y_2)$
    \end{tabular}\\
  \begin{tabular}{l}
    \verb|line|$(x_1,y_1,x_2,y_2,$\\
    \hspace{1cm}$\text{arrow}\in\{0,1\},\text{dashed}\in\{0,1\})$
  \end{tabular}&\begin{tabular}{l}
    Line from $(x_1,y_1)$ to  $(x_2,y_2)$,\\\hspace{1cm}optionally with an arrow and/or dashed
    \end{tabular}\\
  \begin{tabular}{l}
    \verb|STOP|
  \end{tabular}&\begin{tabular}{l}
    Finishes spec inference
    \end{tabular}
\\  \bottomrule
\end{tabular}
\end{table}

For the model in Fig.~\ref{architecture}, the distribution over the next drawing command factorizes as:
\begin{equation}
  \probability_\theta [t_1t_2\cdots t_K | I,S] = \prod_{k = 1}^K \probability_\theta \left[t_k | a_\theta \left(f_\theta(I,\text{render}(S)) | \{t_j\}_{j = 1}^{k - 1}\right), \{t_j\}_{j = 1}^{k - 1}\right]
\end{equation}
where $t_1t_2\cdots t_K$ are the tokens in the drawing command, $I$ is
the target image, $S$ is a spec, $\theta$ are the
parameters of the neural network, $f_\theta(\cdot,\cdot)$ is the
image feature extractor (convolutional network), and $a_\theta(\cdot|\cdot)$ is an attention mechanism. The distribution over
specs factorizes as:
\begin{equation}
  \probability_\theta [S|I] = \prod_{n = 1}^{|S|} \probability_\theta [S_n | I,S_{1:(n-1)}]\times\probability_\theta [\verb|STOP| | I,S]\label{objective}
\end{equation}
where $|S|$ is the length of spec $S$, the subscripts
on $S$ index drawing commands within the spec (so $S_n$ is a sequence of tokens: $t_1t_2\cdots t_K$), and the \verb|STOP|
token is emitted by the network to signal that the spec
explains the image.
We trained our network by sampling specs $S$ and target
images $I$ for randomly generated scenes\footnote{Because the rendering process ignores the ordering of drawing commands in the spec, the mapping from spec to image is many-to-one. When generating random training data for the neural network, we put the drawing commands into a canonical order (left-to-right, top-to-bottom, first drawing circles, then rectangles, and finally lines/arrows.)}
and maximizing $\probability_\theta[S|I]$,
 the likelihood of $S$ given $I$, with respect to
  model parameters $\theta$, by gradient ascent.
We trained on $10^5$ scenes, which takes a day on an Nvidia TitanX GPU.


Our network can ``derender'' random synthetic images
by doing a beam search to
recover  specs maximizing $\probability_\theta[S|I]$. 
 But, if the network predicts an incorrect
drawing command, it has no way of recovering from that error.  
For added robustness 
we treat the
 network outputs as proposals for a Sequential Monte Carlo (SMC) sampling scheme~\citep{SMCBook}.
Our SMC sampler draws samples
from the distribution $\propto L(I|\text{render}(S))
\probability_\theta[S|I]$, where $L(\cdot | \cdot)$
uses the pixel-wise distance between two images as a proxy for a
likelihood.
Here, the network is learning a proposal distribution to amortize the cost of inverting a generative model (the renderer)~\citep{paige2016inference}.
Unconventionally, the target distribution of the SMC sampler
includes the likelihood under the proposal distribution.
Intuitively, both the proposal
distribution and the distance function offer complementary signals for
whether a drawing command is correct,
and we found that combining these signals gave higher accuracy.

\begin{figure}[h]\centering
  \begin{minipage}[c]{7cm}
      \includegraphics[width = 7cm]{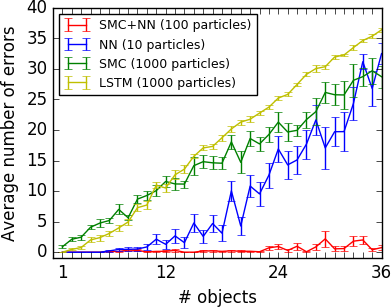}    
  \end{minipage}\hspace{0.5cm}%
  \begin{minipage}[c]{6cm}
      \caption{Parsing~\LaTeX~output after training on diagrams with $\leq 12$ objects. Out-of-sample generalization: Model generalizes to scenes with many more objects ($\approx$ at ceiling when tested on twice as many objects as were in the training data). Neither SMC nor the neural network are sufficient on their own. \# particles varies by model: we compare the models \emph{with equal runtime} ($\approx 1$ sec/object). Average number of errors is (\# incorrect drawing commands predicted by model)$+$(\# correct commands that were not predicted by model).}\label{syntheticResults}
    \end{minipage}
  \end{figure}

\textbf{Experiment 1: Figure~\ref{syntheticResults}.}
  To evaluate which components of the model are necessary to parse complicated scenes,
  we compared  the neural network
  with SMC against the neural network by
itself (using beam search) or SMC by itself.  Only the combination of the two passes a
critical test of generalization: when trained on images with $\leq 12$
objects, it successfully parses scenes with many more objects than the
training data.
We also compare with a baseline that produces the spec in one shot by
using the CNN to extract features of the input which are passed to an LSTM which finally predicts
the spec token-by-token (LSTM in Fig.~\ref{syntheticResults}; Appendix~\ref{captioningBaseline}).
This architecture is used in several successful neural models of image captioning (e.g.,~\cite{vinyals2015show}),
but, for this domain, cannot parse cluttered scenes with many objects.

\subsection{Generalizing to real hand drawings}\label{generalizingTheHandDrawings}
\begin{wrapfigure}{R}{6cm}
\vspace{-0.7cm}\centering\vspace{-0.7cm}  \begin{minipage}[t]{2.5cm}\centering\includegraphics[width = 2cm]{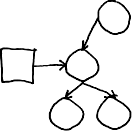}
    \subcaption{hand drawing}
  \end{minipage}\\
   \begin{minipage}[t]{2.5cm}\includegraphics[width = 2cm]{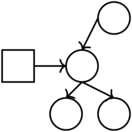}
    \subcaption{rendering of (a)'s inferred spec}
  \end{minipage}\qquad%
  \begin{minipage}[t]{2.5cm}\includegraphics[width = 2cm]{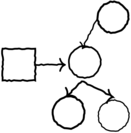}
    \subcaption{noisy rendering of (b)}
  \end{minipage}%
  \caption{Noisy renderings produced in \LaTeX~Ti\emph{k}Z w/ \texttt{pencildraw} package (Appendix~\ref{noisyAppendix})}
    \label{handDrawingExamples} \vspace{-1cm}
\end{wrapfigure}

 We trained the model
to generalize to hand drawings by introducing noise into the
renderings of the training target images, where the noise process  mimics the kinds of variations found in hand drawings (Fig.~\ref{handDrawingExamples}).
While our neurally-guided SMC procedure
used pixel-wise distance as a surrogate for a likelihood function ($L(\cdot|\cdot)$ in Sec.~\ref{neuralNetworkSection}),
 pixel-wise distance fares poorly on hand drawings, which never exactly match
the model's renders.
So, for hand drawings,
we learn a surrogate likelihood function,
$L_{\text{learned}}(\cdot|\cdot)$.
The density $L_{\text{learned}}(\cdot|\cdot)$ is predicted by a convolutional network that we train to predict
the distance between two specs conditioned upon their renderings.
We train $L_{\text{learned}}(\cdot |\cdot )$  to approximate the symmetric difference,
which is  the number of drawing commands by which two specs  differ:
  \begin{equation}
    -\log L_{\text{learned}}(\text{render}(S_1)|\text{render}(S_2))\approx |S_1 - S_2| + |S_2 - S_1|\label{symmetricDistance}
  \end{equation}
  Appendix~\ref{distanceAppendix} details the architecture and training of $L_{\text{learned}}$.

\begin{figure}[H]
  \begin{minipage}[t]{8.25cm}
    \begin{tabular}{llll}
      \includegraphics[width = 2cm]{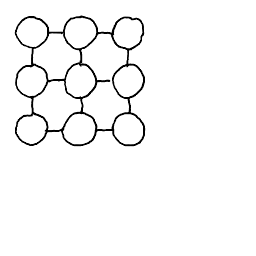}&
      \includegraphics[width = 2cm]{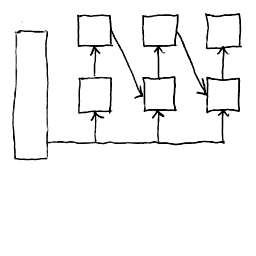}&
      \includegraphics[width = 2cm]{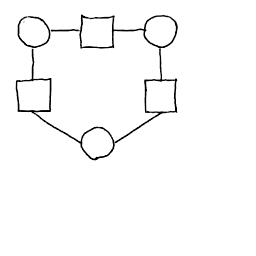}&
      \includegraphics[width = 2cm]{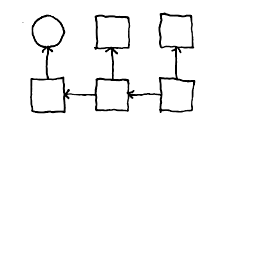}
      \\
      \includegraphics[width = 2cm]{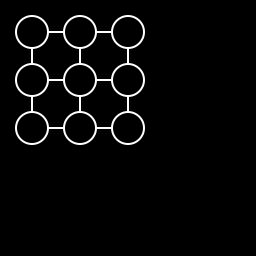}     &
      \includegraphics[width = 2cm]{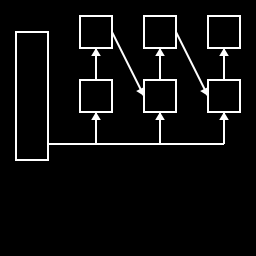}     &
      \includegraphics[width = 2cm]{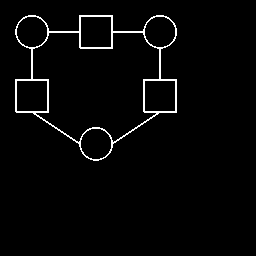}    &
      \includegraphics[width = 2cm]{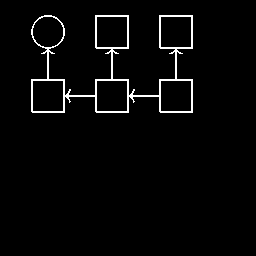}    
      \end{tabular}
    \caption{Left to right: Ising model, recurrent network architecture, figure from a deep learning textbook \cite{Goodfellow-et-al-2016}, graphical model}\label{drawingSuccesses}
  \end{minipage}
\hfill  \begin{minipage}[t]{4.25cm}
    \begin{tabular}{ll}
          \includegraphics[width = 2cm]{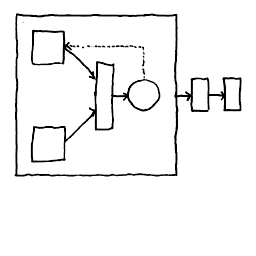}&
          \includegraphics[width = 2cm]{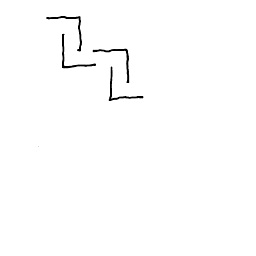}\\
      \includegraphics[width = 2cm]{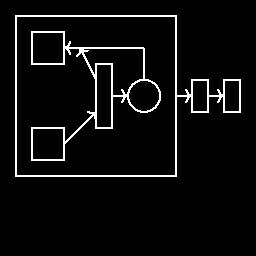}     &
              \includegraphics[width = 2cm]{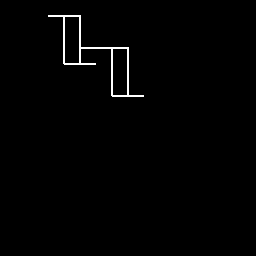}    
    \end{tabular}

    \caption{Near misses. Rightmost: illusory contours (note: no SMC in rightmost)}\label{drawingFailures}
  \end{minipage}

\end{figure}
\begin{figure}[h]\centering
  \begin{minipage}[c]{0.57\textwidth} 
    \centering  \includegraphics[width = 8cm]{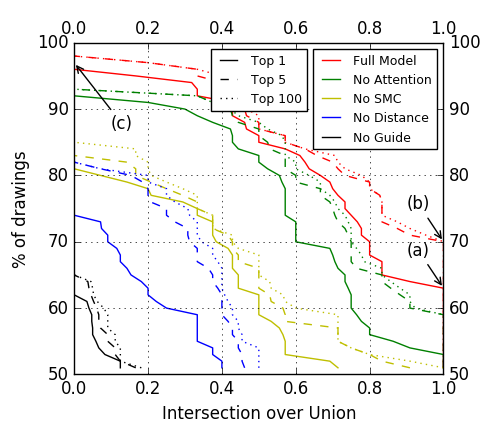}            \vspace{-0.5cm} 
  \end{minipage}\hfill%
      \begin{minipage}[c]{0.4\textwidth} 
  \caption{How close are the model's outputs to the ground truth on hand drawings, as we consider larger sets of samples (1, 5, 100)?
    Distance to ground truth measured by the intersection over union (IoU) of predicted vs. ground truth: IoU of sets $A$ and $B$ is $|A\cap B|/|A\cup B|$. (a) for 63\% of drawings the model's top prediction is exactly correct; (b) for 70\% of drawings the ground truth is in the top 5 model predictions; (c) for 4\% of drawings all of the model outputs have no overlap with the ground truth. Red: the full model. Other colors: lesioned versions of our model.}\label{drawingIntersectionOverUnion}            \vspace{-0.5cm}
      \end{minipage}

\end{figure}

  \textbf{Experiment 2: Figures~\ref{drawingSuccesses}--\ref{drawingIntersectionOverUnion}.}
    We evaluated, but did not train, our system on 100 real hand-drawn figures; see Fig.~\ref{drawingSuccesses}--\ref{drawingFailures}.
    These were drawn carefully but not perfectly with the aid of graph paper.
    For each drawing we annotated a ground truth spec and had the neurally guided SMC sampler
produce $10^3$ samples. 
For 63\% of the drawings, the Top-1 most likely sample exactly matches the
ground truth; with more samples, the model finds specs
that are closer to the ground truth annotation (Fig.~\ref{drawingIntersectionOverUnion}).
We will show that the program synthesizer
corrects some of these small errors (Sec.~\ref{synthesizerHelpsParsing}).
Because the model sometimes makes mistakes on hand drawings,
we envision it working as follows:
a user sketches a diagram,
and the system responds by proposing a few candidate interpretations.
The user could then select the one closest to their intention and edit it if necessary.

\setlength\tabcolsep{3.5pt}
\setlength\tabcolsep{6pt}

\section{Synthesizing graphics programs from specs}\label{programSynthesisSection}
Although the spec describes the contents
of a scene, it does not encode higher-level features of an image
such as repeated motifs or symmetries, which are more naturally captured by a graphics program.
We seek to synthesize graphics programs from their specs.

Although it might seem desirable to synthesize programs in a Turing-complete language such as Lisp or Python, a more tractable approach is to specify
what in the program
languages community is called a Domain Specific Language (DSL)~\citep{polozov2015flashmeta}. Our DSL (Tbl.~\ref{DSL})
encodes prior knowledge of what graphics programs tend to look like.

\begin{table}[H]
  \caption{Grammar over graphics programs. We allow loops (\texttt{for}) with conditionals (\texttt{if}), vertical/horizontal reflections (\texttt{reflect}), variables (Var) and affine transformations ($\mathbb{Z}\times$Var\texttt{+}$\mathbb{Z}$).}\label{DSL}
  \begin{tabular}{rl}\toprule
  Program$\to$&Statement; $\cdots$; Statement\\
  Statement$\to$&\texttt{circle}(Expression,Expression)\\
  Statement$\to$&\texttt{rectangle}(Expression,Expression,Expression,Expression)\\
  Statement$\to$&\texttt{line}(Expression,Expression,Expression,Expression,Boolean,Boolean)\\
  Statement$\to$&\texttt{for}$(0\leq \text{Var}  < \text{Expression})$\texttt{ \{ if }$(\text{Var} > 0)$\texttt{ \{ }Program\texttt{ \}; }Program\texttt{ \}}\\
  Statement$\to$&\texttt{reflect}$(\text{Axis})$\texttt{ \{ }Program\texttt{ \}}\\
  Expression$\to$&$\mathbb{Z}\times$Var\texttt{+}$\mathbb{Z}$\\
  Axis$\to$&\texttt{X = }$\mathbb{Z}$ | \texttt{Y = }$\mathbb{Z}$\\
    $\mathbb{Z}\to$&an integer\\\bottomrule
  \end{tabular}
\end{table}

Given the DSL and a spec $S$, we want a program that both satisfies $S$
and, at the same time, is the ``best'' explanation of $S$.
For example, we might prefer more general programs or, in the spirit of Occam's razor,
prefer shorter programs.
We wrap these intuitions up into a cost function over programs,
and seek the minimum cost program consistent with $S$:
\begin{equation}
    \text{program}(S) = \argmax_{p\in \text{DSL}} \indicator\left[p \text{ consistent w/ } S \right]\exp \left( -\text{cost}(p) \right)\label{programObjective}
\end{equation}
We define the
cost of a program to be the number of Statement's it contains (Tbl.~\ref{DSL}).
We also penalize using many different numerical constants; see Appendix~\ref{costAppendix}.
Returning to the generative model in Fig.~\ref{roadmap},
this setup is the same as saying that the prior probability of a program $p$ is $\propto \exp\left(-\text{cost}(p) \right)$ and the likelihood of a spec $S$ given a program $p$ is $\indicator[p\text{ consistent w/ }S]$.

The constrained optimization problem in
Eq.~\ref{programObjective} is intractable in general, but there
exist efficient-in-practice tools for finding exact solutions to such
program synthesis problems. We use the state-of-the-art Sketch
tool~\citep{solar2008program}.
Sketch takes as input a space of programs, along with
a specification of the program's behavior and optionally a cost
function.  It translates the synthesis problem into a constraint
satisfaction problem and then uses a SAT solver to find a minimum-cost
program satisfying the specification.  Sketch requires a
 \emph{finite program space}, which here means that the depth of the
program syntax tree is bounded (we set the bound to 3),
but has the guarantee that it 
always eventually finds a globally optimal solution.
In exchange for this optimality guarantee
it comes with no guarantees
on runtime.
For our domain synthesis times vary from minutes to hours,
with 27\% of the drawings timing out the synthesizer after 1 hour.
Tbl.~\ref{exampleSynthesisResults} shows programs recovered by our system.
A main impediment to our use of these general techniques is
the prohibitively high cost of searching for programs.
We next describe how to learn to synthesize programs much faster (Sec.~\ref{learningASearchPolicy}),
timing out on 2\% of the drawings and solving 58\% of problems within a minute.

\newcommand{\exampleProgramSize}{4cm}
\newcommand{\exampleTraceSize}{3.5cm}
\newcommand{\exampleDrawingSize}{1.25cm}
\lstset{basicstyle = \scriptsize\ttfamily}

\begin{table}[t]
  \caption{Drawings (left), their specs (middle left), and programs synthesized from those specs (middle right). Compared to the specs the programs are more compressive (right: programs have fewer lines than specs) and automatically group together related drawing commands. Note the nested loops  and conditionals in the Ising model, combination of symmetry and iteration in the bottom figure,  affine transformations in the top figure, and the complicated program in the second figure to bottom.}\label{exampleSynthesisResults}
\centering  \begin{tabular}{m{1.5cm}llc}
    \toprule
    \textbf{Drawing}&\textbf{Spec}&\textbf{Program}&
      \textbf{Compression factor}
      \\
    \midrule
    \includegraphics[width = \exampleDrawingSize]{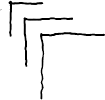}&
\begin{minipage}{\exampleTraceSize}\begin{lstlisting}
Line(2,15, 4,15)
Line(4,9, 4,13)
Line(3,11, 3,14)
Line(2,13, 2,15)
Line(3,14, 6,14)
Line(4,13, 8,13)
\end{lstlisting}
\end{minipage}&     \begin{minipage}{\exampleProgramSize} \begin{lstlisting}
for(i<3)
 line(i,-1*i+6,
      2*i+2,-1*i+6)
 line(i,-2*i+4,i,-1*i+6)
       \end{lstlisting}
     \end{minipage}&$\frac{6}{3} = 2\text{x}$\\\midrule
     \includegraphics[width = \exampleDrawingSize]{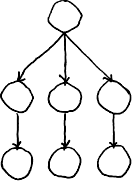}&
\begin{minipage}{\exampleTraceSize}\begin{lstlisting}
Line(5,13,2,10,arrow)
Circle(5,9)
Circle(8,5)
Line(2,8, 2,6,arrow)
Circle(2,5)
\end{lstlisting}
\small\emph{... etc. ...; 13 lines}
\end{minipage}&
             \begin{minipage}{\exampleProgramSize}\begin{lstlisting}
circle(4,10)
for(i<3)
 circle(-3*i+7,5)
 circle(-3*i+7,1)
 line(-3*i+7,4,-3*i+7,2,arrow)
 line(4,9,-3*i+7,6,arrow)
\end{lstlisting}
\end{minipage}&$\frac{13}{6} = 2.2\text{x}$\\\midrule
    \includegraphics[width = \exampleDrawingSize]{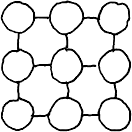}&
\begin{minipage}{\exampleTraceSize}\begin{lstlisting}
Circle(5,8)
Circle(2,8)
Circle(8,11)
Line(2,9, 2,10)
Circle(8,8)
Line(3,8, 4,8)
Line(3,11, 4,11)
\end{lstlisting}
  \small\emph{... etc. ...; 21 lines}
\end{minipage}&\begin{minipage}{\exampleProgramSize}
\begin{lstlisting}
for(i<3)
 for(j<3)
  if(j>0)
   line(-3*j+8,-3*i+7,
        -3*j+9,-3*i+7)
   line(-3*i+7,-3*j+8,
        -3*i+7,-3*j+9)
  circle(-3*j+7,-3*i+7)
\end{lstlisting}
\end{minipage}&$\frac{21}{6} = 3.5\text{x}$\\\midrule
\includegraphics[width = \exampleDrawingSize]{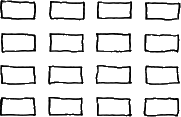}&
\begin{minipage}{\exampleTraceSize}\begin{lstlisting}
Rectangle(1,10,3,11)
Rectangle(1,12,3,13)
Rectangle(4,8,6,9)
Rectangle(4,10,6,11)
\end{lstlisting}
  \small\emph{... etc. ...; 16 lines}
\end{minipage}&\begin{minipage}{\exampleProgramSize}
\begin{lstlisting}
for(i<4)
 for(j<4)
  rectangle(-3*i+9,-2*j+6,
            -3*i+11,-2*j+7)
\end{lstlisting}
\end{minipage}&$\frac{16}{3} = 5.3\text{x}$\\\midrule

  \includegraphics[width = \exampleDrawingSize]{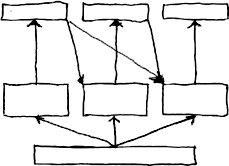}&
\begin{minipage}{\exampleTraceSize}\begin{lstlisting}
Line(3,10,3,14,arrow)
Rectangle(11,8,15,10)
Rectangle(11,14,15,15)
Line(13,10,13,14,arrow)
  \end{lstlisting}\small\emph{... etc. ...; 16 lines}%
  \end{minipage}&\begin{minipage}{\exampleProgramSize}
\begin{lstlisting}
for(i<3)
 line(7,1,5*i+2,3,arrow)
 for(j<i+1)
  if(j>0)
   line(5*j-1,9,5*i,5,arrow)
  line(5*j+2,5,5*j+2,9,arrow)
 rectangle(5*i,3,5*i+4,5)
 rectangle(5*i,9,5*i+4,10)
rectangle(2,0,12,1)
\end{lstlisting}
\end{minipage}&$\frac{16}{9} = 1.8\text{x}$\\\midrule    

  \includegraphics[width = \exampleDrawingSize]{figures/expert-72-trim.png}&

\begin{minipage}{\exampleTraceSize}\begin{lstlisting}
Circle(2,8)
Rectangle(6,9, 7,10)
Circle(8,8)
Rectangle(6,12, 7,13)
Rectangle(3,9, 4,10)
\end{lstlisting}\small\emph{... etc. ...; 9 lines}
  \end{minipage}&\begin{minipage}{\exampleProgramSize}
\begin{lstlisting}
reflect(y=8)
 for(i<3)
  if(i>0)
   rectangle(3*i-1,2,3*i,3)
  circle(3*i+1,3*i+1)
\end{lstlisting}
\end{minipage}&$\frac{9}{5} = 1.8\text{x}$ \\\bottomrule
  \end{tabular}
  \end{table}

\subsection{Learning a search policy for synthesizing programs}\label{learningASearchPolicy}

We want to leverage powerful, domain-general techniques from the program synthesis community,
but make them much faster by
learning a domain-specific \textbf{search policy}.
A search policy poses search problems
like those in Eq.~\ref{programObjective},
but also offers additional constraints on the structure of the program (Tbl.~\ref{policyOutput}).
For example, a policy might decide to first try searching over small programs before searching over large programs,
or decide to prioritize searching over programs that have loops.

Formally, our search policy, $\pi_\theta(\sigma  | S )$, takes as input a spec $S$ and predicts a distribution over search problems, each of which is written $\sigma $ and corresponds to a set of possible programs to search over (so $\sigma \subseteq \text{DSL}$).
We assume a finite\footnote{It is not strictly necessary that $\Sigma$ be a finite set, only that it be recursively enumerable.
For example, Levin Search considers the setting where the infinite set of all Turing machines serves as $\Sigma$.} family of
search problems, which we write $\Sigma$,
and require that every program in the DSL
is contained in at least one $\sigma \in \Sigma$.

Good policies will prefer tractable program spaces,
so that the search procedure will terminate early, 
but should also prefer program spaces likely to contain
programs that concisely explain the data.
These two desiderata are in tension:
tractable synthesis problems involve searching over smaller spaces,
but smaller spaces are less likely to contain good programs.
Our goal now is to find the parameters of the policy, written $\theta$, that best navigate this trade-off.

Given a search policy, what is the best way of using it to quickly find minimum cost programs?
We use a bias-optimal search algorithm (c.f. Schmidhuber 2004~\citep{schmidhuber2004optimal}):

\noindent\textbf{Definition: Bias-optimality.} 
   A search algorithm is $n$-\emph{bias optimal}
with respect to a distribution $\probability_{\text{bias}}[\cdot ]$ if it is
guaranteed to find a solution in $\sigma $ after searching for at least time
$n\times\frac{t(\sigma )}{\probability_{\text{bias}}[\sigma ]}$, where $t(\sigma )$ is the time it
takes to verify that $\sigma $ contains a solution to the
search problem.

Bias-optimal search over program spaces is known as \textbf{Levin Search}~\cite{levin1973universal}; an example of a $1$-bias optimal search algorithm is an ideal time-sharing system that allocates $\probability_{\text{bias}}[\sigma ]$ of its time to trying $\sigma $.  We construct a $1$-bias optimal search algorithm by identifying $\probability_{\text{bias}}[\sigma ] = \pi_\theta(\sigma |S)$ and $t(\sigma )  = t(\sigma|S)$, where $t(\sigma|S)$ is how long the synthesizer takes to search $\sigma $ for a program for $S$. Intuitively, this means that the search algorithm explores the entire program space, but spends most of its time in the regions of the space that  the policy judges to be most promising.
Concretely,
this means that our
synthesis strategy is to
run many different
program searches \emph{in parallel} (i.e., run in parallel different instances of the synthesizer, one for each  $\sigma\in \Sigma$),
but to allocate
compute time to a $\sigma $ in proportion to $\pi_\theta(\sigma |S)$.

\begin{figure}\centering
  \begin{tikzpicture}
    \node[anchor = west] at (0,5.25) {Entire program search space};
    \draw[fill = yellow,fill opacity = 0.15,draw = black] (0,0) rectangle (5,5);
    \draw [fill = yellow, opacity = 0.2] (0,0)--(0,2)--(4,0)--(0,0);
    \draw [fill = yellow, opacity = 0.4] (0,5)--(0,1)--(3,5);
    \draw [fill = yellow, opacity = 0.6] (2.5,5)--(3.2,0)--(5,0)--(5,5);
    \draw (0,2) -- node[below,sloped]{short programs} (4,0);
    \draw (0,2) -- node[above,sloped]{long programs} (4,0);
    \draw (2.5,5) -- node[above,sloped]{programs w/ reflections} (3.2,0);
    \draw (0,1) -- node[above,sloped]{ programs w/ loops} (3,5);
    \node(p1)[anchor =west ] at (5.5,3.5) {$\pi_\theta(\text{short, no loop/reflect}|S) = $};
    \draw [fill = yellow, fill opacity = 0.2,draw = black]  ([xshift = 0cm,yshift = -0.2cm]p1.east) rectangle ([xshift = 0.4cm,yshift = 0.2cm]p1.east);
    \node(p2)[anchor =west ] at ([yshift = -0.5cm]p1.west) {$\pi_\theta(\text{long, loops}|S) = $};
    \draw [fill = yellow, fill opacity = 0.4,draw = black]  ([xshift = 0cm,yshift = -0.2cm]p2.east) rectangle ([xshift = 0.4cm,yshift = 0.2cm]p2.east);
    \node(p3)[anchor =west ] at ([yshift = -0.5cm]p2.west) {$\pi_\theta(\text{long, no loop/reflect}|S) = $};
    \draw [fill = yellow, fill opacity = 0.15,draw = black]  ([xshift = 0cm,yshift = -0.2cm]p3.east) rectangle ([xshift = 0.4cm,yshift = 0.2cm]p3.east);
    \node(p4)[anchor =west ] at ([yshift = -0.5cm]p3.west) {$\pi_\theta(\text{long, reflects}|S) = $};
    \draw [fill = yellow, fill opacity = 0.6,draw = black]  ([xshift = 0cm,yshift = -0.2cm]p4.east) rectangle ([xshift = 0.4cm,yshift = 0.2cm]p4.east);
    \node(p5)[anchor =west ] at ([yshift = -0.5cm]p4.west) {\emph{etc.}};
  \end{tikzpicture}
  \caption{The bias-optimal search algorithm divides the entire (intractable) program search space in to (tractable) program subspaces (written $\sigma $), each of which contains a restricted set of programs. For example, one subspace might be short programs which don't loop. The policy $\pi$ predicts a distribution over program subspaces. The weight that $\pi$ assigns to a subspace is indicated by its yellow shading in the above figure, and is conditioned on the spec $S$.}
  \end{figure}
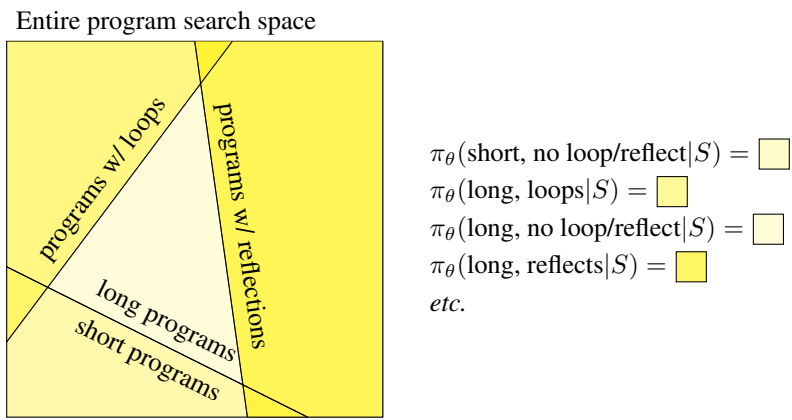

Now in theory any $\pi_\theta(\cdot |\cdot ) $ is a bias-optimal searcher.
But the actual runtime of the algorithm depends strongly upon
the bias $\probability_{\text{bias}}[\cdot ]$.
Our new approach is to learn $\probability_{\text{bias}}[\cdot ]$
by picking the policy minimizing the
expected bias-optimal time to solve a training corpus, $\mathcal{D}$, of graphics program synthesis problems:\footnote{This loss is differentiable but nonconvex even if $\pi_{\theta}(\cdot |\cdot )^{-1}$ is convex.}
\begin{align}
\textsc{Loss}(\theta ; \mathcal{D})& =  \expect_{S\sim\mathcal{D}}\left[ \min_{\sigma\in \text{\textsc{Best}}(S)}\frac{t(\sigma | S)}{\pi_\theta (\sigma | S)}\right] + \lambda \Vert\theta\Vert_2^2\label{policyLoss}\\
\text{where }  \sigma \in\text{\textsc{Best}}(S) &\text{ if  a minimum cost program for }S \text{ is in }\sigma .\nonumber 
\end{align}

To generate a training corpus for learning a policy,
we synthesized minimum cost programs for each of our hand drawings
and for each $\sigma $,
then minimized Eq.~\ref{policyLoss} using gradient descent while annealing a softened minimum to the hard minimization in Eq.~\ref{policyLoss} (see Appendix~\ref{policyAppendix}).
Because we want to learn a policy from only $100$ drawings,
we parameterize $\pi$ with a  low-capacity bilinear model:
\begin{equation}
  \pi_{\theta}(\sigma |S)\propto \exp \left( \phi_{\text{params}}(\sigma )^\top\theta \phi_{\text{spec}}(S)\right)
\end{equation}
where $\phi_{\text{params}}(\sigma  )$ is a one-hot encoding of
the parameter settings of $\sigma $ (see Tbl.~\ref{policyOutput})
and $\phi_{\text{spec}}(S)$ extracts a vector of counts of the drawing primitives in $S$;
thus $\theta$ has only 96 real-valued parameters.\footnote{$\theta$ has only 96 parameters because it
  is a matrix mapping a 4-dimensional feature vector into a 24-dimensional output space.
  The output space is 24-dimensional because $\sigma $ assumes one of 24 different values,
  and the input space is 4-dimensional because we have three different drawing primitives,
along with an extra dimension for a `bias' term.}

\textbf{Experiment 3: Table~\ref{policyEvaluation}; Figure~\ref{policyHistogram}.}
We compare synthesis times for our learned search policy
with 4 alternatives:
 \emph{Sketch}, which poses the
 entire problem wholesale to the Sketch program synthesizer;
 \emph{DC}, a DeepCoder--style model that learns to predict which program components (loops, reflections)
 are likely to be useful~\cite{BalGauBroetal16} (Appendix~\ref{dcAppendix});
 \emph{End--to-End},
 which trains a recurrent neural network to regress directly from images to programs (Appendix~\ref{eeAppendix});
and an \emph{Oracle},
a policy which always picks the quickest to search $\sigma $
 also containing a minimum cost program.
Our approach improves upon Sketch by itself,
and comes close to the Oracle's performance.
One could never construct this Oracle,
because the agent does not know ahead of time which
$\sigma $'s contain minimum cost programs nor does it know how long each
$\sigma $ will take to search.
With this learned policy in hand we can synthesize 58\% of programs within a minute.
\begin{table}[h]\centering
  \caption{Parameterization of different ways of posing the program synthesis problem. The policy learns to choose parameters likely to quickly yield a minimal cost program.
  }\label{policyOutput}
  \begin{tabular}{lll}\toprule
  Parameter&Description&Range\\\midrule
  Loops?&Is the program allowed to loop?&$\{\text{True},\text{False}\}$\\
  Reflects?&Is the program allowed to have reflections?&$\{\text{True},\text{False}\}$\\
  Incremental?&Solve the problem piece-by-piece or all at once?&$\{\text{True},\text{False}\}$\\
  Maximum depth& Bound on the depth of the program syntax tree&$\{1,2,3\}$
  \\\bottomrule
  \end{tabular}\vspace{0.5cm}
  \end{table}
\begin{table}[h]
  \centering
  \begin{minipage}[c]{0.4\textwidth}
  \begin{tabular}{lll}
    \toprule Model&\begin{tabular}{c}
      Median \\search time
    \end{tabular}&\begin{tabular}{c}
      Timeouts\\(1 hr)
      \end{tabular}\\\midrule
    Sketch&274 sec&27\%\\
    DC&187 sec&2\%\\
    End--to--End&63 sec&94\%\\
    Oracle&6 sec&2\%\\
    Ours&28 sec&2\%\\\bottomrule 
    \end{tabular}
  \end{minipage}\hfill%
  \begin{minipage}[c]{0.5\textwidth}
    \caption{Time to synthesize a minimum cost program. Sketch: out-of-the-box performance of Sketch~\citep{solar2008program}. DC: Deep--Coder style baseline that predicts program components, trained like~\cite{BalGauBroetal16}. End--to--End: neural net trained to regress directly from images to programs, which fails to find valid programs 94\% of the time. Oracle:  upper bounds  the performance of any bias--optimal search policy. Ours:  
           evaluated w/ 20-fold cross validation.}\label{policyEvaluation}
  \end{minipage}

\vspace{1cm}
  
  \includegraphics[width = \textwidth]{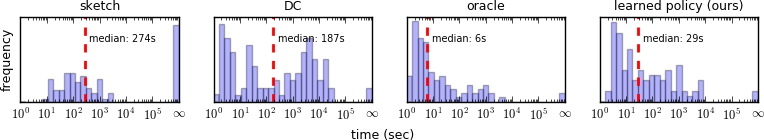}%
  \captionof{figure}{Time to synthesize a minimum cost program (compare w/ Table~\ref{policyEvaluation}). End--to--End: not shown because it times out on 96\% of drawings, and has its median time (63s) calculated only on non-timeouts, wheras the other comparisons include timeouts in their median calculation. $\infty = $ timeout. Red dashed line is median time.}\label{policyHistogram}
\end{table}

\section{Applications of graphics program synthesis}\label{applicationsSection}

\subsection{Correcting errors made by the neural network}\label{synthesizerHelpsParsing}
The program synthesizer can help correct errors from the execution spec proposal network by favoring  specs which lead to more concise or general programs.
For example, one generally prefers figures with perfectly aligned objects over figures whose parts are slightly misaligned -- and precise alignment lends itself to short programs.
Similarly, figures often have repeated parts,
which the program synthesizer might be able to model as a loop or reflectional symmetry.
So, in considering several candidate specs proposed by the neural network,
we might prefer specs whose best programs have desirable features such being short or having iterated structures.
Intuitively,
this is like the `top down' influence of cognition upon perception:
a reasoning engine (the program synthesizer)
can influence the agent's percept
through higher-level considerations  like symmetry and alignment.

Concretely, we implemented the following scheme: for an image $I$, the neurally guided sampling scheme (Section~\ref{neuralNetworkSection}) samples a set of candidate specs, written $\mathcal{F}(I)$.
Instead of predicting the most likely spec in $\mathcal{F}(I)$ according to the neural network,
we can take into account the programs that best explain the specs. 
Writing $\hat{S}(I)$ for the spec the model predicts for image $I$,
\begin{equation}
\hat{S}(I) = \argmax_{S\in \mathcal{F}(I)} L_{\text{learned}}(I | \text{render}(S))\times \probability_\theta[S|I] \times\probability_{\beta} [ \text{program}(S)] 
\end{equation}
where $\probability_{\beta} [\cdot]$ is a prior probability
distribution over programs parameterized by $\beta$.
This is equivalent to doing
MAP inference in a generative model where the program is first drawn
from $\probability_{\beta} [\cdot]$, then the program is executed deterministically,
and then we observe a noisy version of the program's output, where $L_\text{learned}(I|\text{render}(\cdot))\times\probability_\theta[\cdot|I]$
is our observation model.

Given a corpus of graphics program synthesis problems with annotated ground truth specs (i.e. $(I,S)$ pairs),
we find a maximum likelihood estimate of $\beta$:
\begin{equation}
  \beta^* = \argmax_{\beta} \expect \left[ \log \frac{\probability_{\beta} [\text{program}(S)]\times L_{\text{learned}}(I|\text{render}(S))\times \probability_\theta[S|I]}{\sum_{S'\in \mathcal{F}(I)} \probability_{\beta} [\text{program}(S')]\times L_{\text{learned}}(I|\text{render}(S'))\times \probability_\theta[S'|I]} \right]
\end{equation}
where the expectation is taken both over the model predictions and the
$(I,S)$ pairs in the training corpus.  We define $\probability_{\beta}
[\cdot]$ to be a log linear distribution $\propto \exp
(\beta\cdot \phi(\text{program}))$, where $\phi(\cdot)$ is a feature
extractor for programs.  We extract a few basic features of a
program, such as its size and how many loops it has, and use these
features to help predict whether a spec is the correct explanation
for an image.
\begin{wrapfigure}{r}{5cm}\vspace{-0.5cm}
  \includegraphics[width = 5cm]{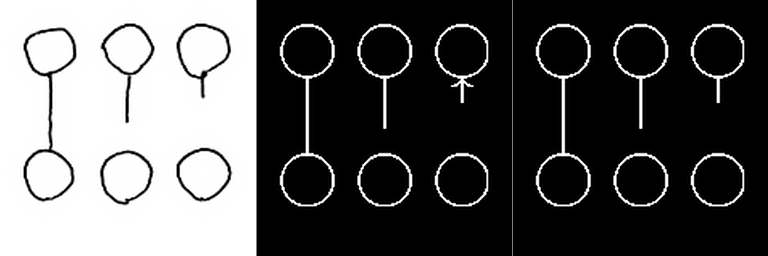}
  \includegraphics[width = 5cm]{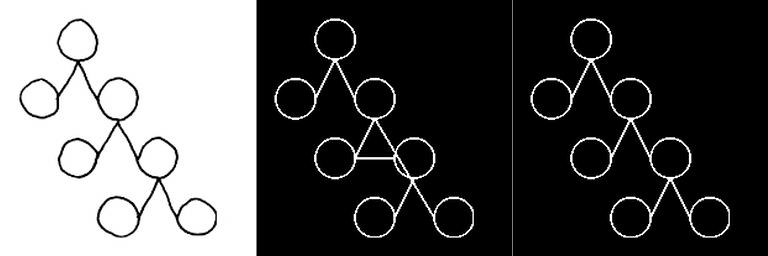}
  \caption{Left: hand drawings. Center: interpretations favored by the deep network. Right: interpretations favored after learning a prior over programs. The prior favors  simpler programs, thus (top) continuing the pattern of not having an arrow is preferred, or (bottom) continuing the ``binary search tree'' is preferred.}\vspace{-1cm}\label{exampleOfProgramCorrectingMistake}
  \end{wrapfigure}
We synthesized programs for the top 10 specs
output by the deep network.  Learning this prior over programs can
help correct mistakes made by the neural network, and also
occasionally introduces mistakes of its own; see
Fig.~\ref{exampleOfProgramCorrectingMistake} for a representative
example of the kinds of corrections that it makes. On the whole
it modestly improves our Top-1 accuracy from 63\% to 67\%.  Recall that
from Fig.~\ref{drawingIntersectionOverUnion} that the best improvement
in accuracy we could possibly get is 70\% by looking at the top 10 specs.



\subsection{Modeling similarity between drawings}
Modeling drawings using programs opens up new ways to measure similarity between them.
For example, we might say that two drawings are similar if they both contain loops of length 4,
or if they share a reflectional symmetry,
or if they are both organized according to a grid-like structure.

We measure the similarity between two drawings by extracting features
of the lowest-cost programs that describe them. Our features are counts of the number of times that different components in the
DSL were used (Tbl.~\ref{DSL}).
We  then find drawings which are either close together or far apart in program feature space.
One could use many
alternative similarity metrics between drawings which would capture pixel-level similarities while missing high-level geometric similarities.
We used our learned distance metric between specs, $L_{\text{learned}}(\cdot|\cdot)$,
to find drawings that are either close together or far apart according to the learned
distance metric over images.
Fig.~\ref{similarityMatrix} illustrates the kinds of drawings that these different metrics put closely together.

\newcommand{\similaritySize}{1cm}
\newcommand{\similarityArrowSize}{0.35cm}
\begin{figure}[t]
  \begin{tabular}{c|c|c|}
    \multicolumn{1}{c}{}    &\multicolumn{1}{c}{\textbf{Close in program space}}&\multicolumn{1}{c}{\textbf{Far apart in program space}}\\\cline{2-3}
    \multicolumn{1}{c|}{}&&\\
    \multicolumn{1}{c|}{    \begin{tabular}{l}\textbf{Close in }\\ \textbf{image space} \end{tabular}}    &
    \begin{tabular}{l}
      \begin{minipage}{\similaritySize}\includegraphics[width = \similaritySize]{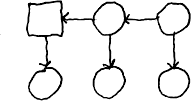}\end{minipage}\begin{minipage}{\similarityArrowSize}$\leftrightarrow$\end{minipage}\begin{minipage}{\similaritySize}\includegraphics[width = \similaritySize]{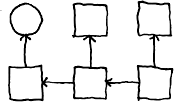}\end{minipage}\\\\
      \begin{minipage}{\similaritySize}\includegraphics[width = \similaritySize]{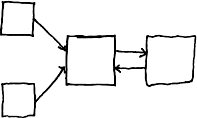}\end{minipage}\begin{minipage}{\similarityArrowSize}$\leftrightarrow$\end{minipage}\begin{minipage}{\similaritySize}\includegraphics[width = \similaritySize]{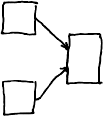}\end{minipage}
    \end{tabular} &
    \begin{tabular}{l}
      \begin{minipage}{\similaritySize}\includegraphics[width = \similaritySize]{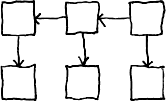}\end{minipage}\begin{minipage}{\similarityArrowSize}$\leftrightarrow$\end{minipage}\begin{minipage}{\similaritySize}\includegraphics[width = \similaritySize]{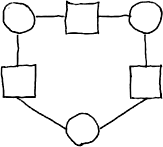}\end{minipage}\\\\
      \begin{minipage}{\similaritySize}\includegraphics[width = \similaritySize]{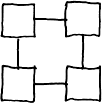}\end{minipage}\begin{minipage}{\similarityArrowSize}$\leftrightarrow$\end{minipage}\begin{minipage}{\similaritySize}\includegraphics[width = \similaritySize]{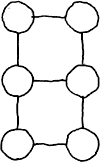}\end{minipage}
    \end{tabular}\\
    \multicolumn{1}{c|}{}&&\\\cline{2-3}
    \multicolumn{1}{c|}{}&&\\
    \multicolumn{1}{c|}{    \begin{tabular}{c}\textbf{Far apart in }\\ \textbf{image space} \end{tabular}}    &        \begin{tabular}{l}
      \begin{minipage}{\similaritySize}\includegraphics[width = \similaritySize]{figures/expert-52-trim.png}\end{minipage}\begin{minipage}{\similarityArrowSize}$\leftrightarrow$\end{minipage}\begin{minipage}{\similaritySize}\includegraphics[width = \similaritySize]{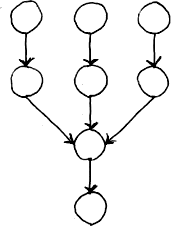}\end{minipage}\\\\
      \begin{minipage}{\similaritySize}\includegraphics[width = \similaritySize]{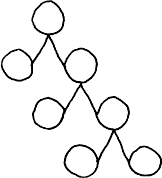}\end{minipage}\begin{minipage}{\similarityArrowSize}$\leftrightarrow$\end{minipage}\begin{minipage}{\similaritySize}\includegraphics[width = \similaritySize]{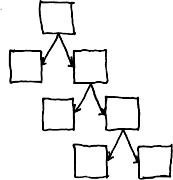}\end{minipage}
    \end{tabular}&
    \begin{tabular}{l}
      \begin{minipage}{\similaritySize}\includegraphics[width = \similaritySize]{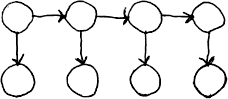}\end{minipage}\begin{minipage}{\similarityArrowSize}$\leftrightarrow$\end{minipage}\begin{minipage}{\similaritySize}\includegraphics[width = \similaritySize]{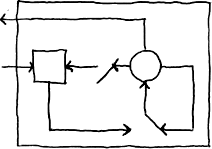}\end{minipage}\\\\
      \begin{minipage}{\similaritySize}\includegraphics[width = \similaritySize]{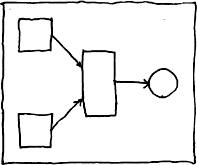}\end{minipage}\begin{minipage}{\similarityArrowSize}$\leftrightarrow$\end{minipage}\begin{minipage}{\similaritySize}\includegraphics[width = \similaritySize]{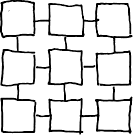}\end{minipage}
    \end{tabular}
\\    \multicolumn{1}{c|}{}&&\\\cline{2-3}\cline{2-3}    
  \end{tabular}
  
  \centering
\caption{Pairs of images either close together or far apart in different features spaces.  The symbol $\leftrightarrow$ points to the compared images. Features of the program capture abstract notions like symmetry and repetition. Distance metric over images is $L_{\text{learned}}(\cdot|\cdot)$ (see Section~\ref{generalizingTheHandDrawings}). The off-diagonal entries highlight the difference between these metrics: similarity of programs captures high-level features like repetition and symmetry, whereas similarity of images correspondends to similar drawing commands being in similar places.}\label{similarityMatrix}  \end{figure}

\subsection{Extrapolating figures}
\begin{figure}[H]\centering
  \includegraphics[width = 0.885\textwidth]{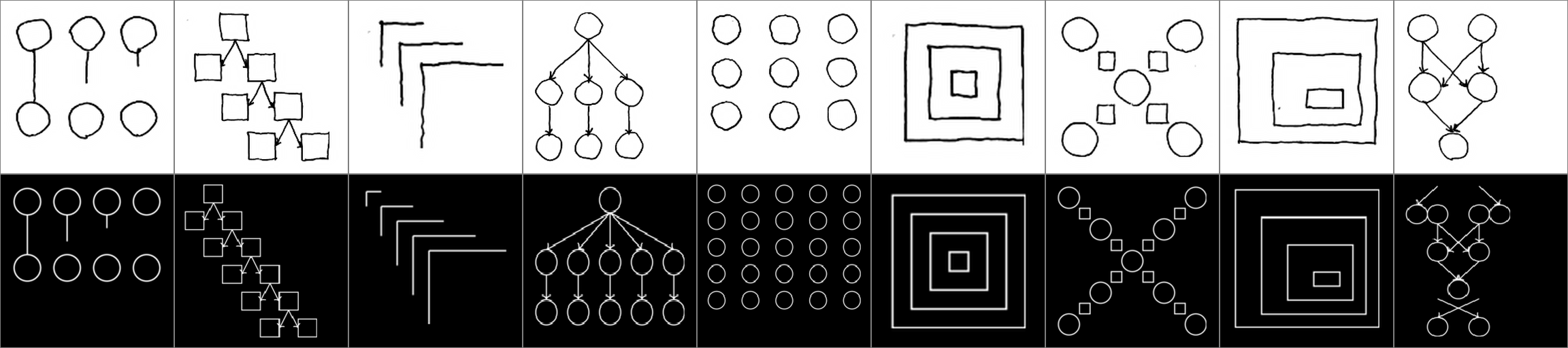}
  \includegraphics[width = 0.885\textwidth]{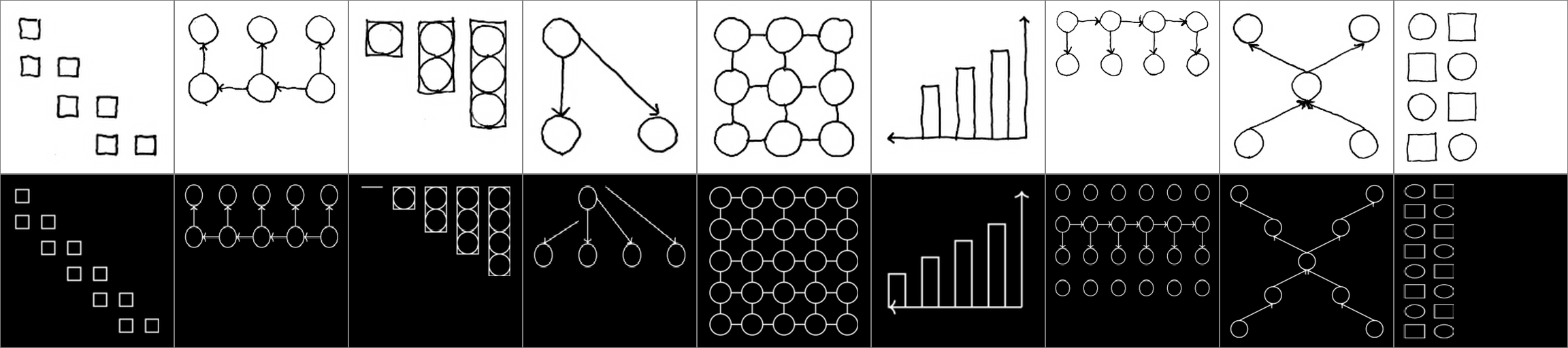}
  \includegraphics[width = 0.885\textwidth]{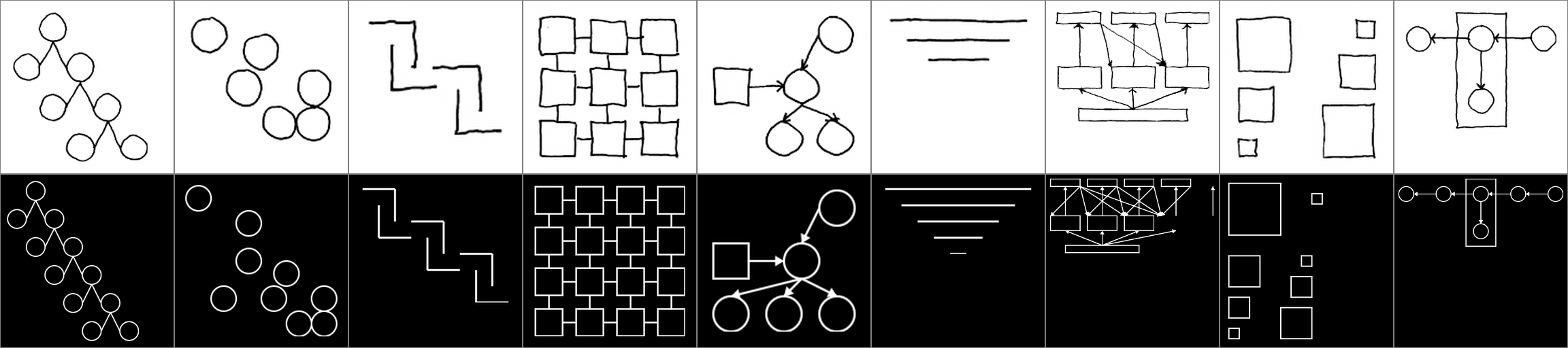}  
   \caption{Top, white: hand drawings. Bottom, black: extrapolations produced by our system.}
  \label{extrapolationFigure}
  \end{figure}

Having access to the source code of a graphics program facilitates coherent, high-level image editing.
For example we can extrapolate figures
by increasing the number of times that loops are executed.
Extrapolating repetitive visuals patterns comes naturally to humans,
and is a practical application:
imagine hand drawing a repetitive graphical model structure
and having our system automatically induce and extend the pattern.
Fig.~\ref{extrapolationFigure} shows extrapolations produced by our system.



\section{Related work}

\textbf{Program Induction:}
Our approach to learning to search for programs draws theoretical
underpinnings from Levin
search~\citep{levin1973universal,solomonoff1984optimum} and
Schmidhuber's OOPS model~\citep{schmidhuber2004optimal}.
DeepCoder~\citep{BalGauBroetal16} is a recent model which, like ours, learns to predict likely program components.
Our work differs by identifying and modeling
the trade-off between tractability and probability of success.
TerpreT~\citep{gaunt2016terpret} 
systematically compares constraint-based program synthesis techniques
against gradient-based search methods, like those used to train
Differentiable Neural Computers~\citep{graves2016hybrid}.  The TerpreT
experiments motivate our use of constraint-based techniques.

\textbf{Deep Learning:} Our neural network combines the architectural ideas of Attend-Infer-Repeat~\citep{eslami1603attend} -- which learns to decompose an image into its constituent objects -- with the training regime and SMC inference  of Neurally Guided Procedural Modeling~\cite{ritchie2016neurally} -- which learns to control procedural graphics programs.
IM2LATEX~\citep{im2latex} is a recent work that 
 derenders  \LaTeX~ equations,
recovering a markup language representation.
Our goal is to go from
noisy input to a high-level program,
which goes beyond markup languages by supporting
programming constructs like loops and conditionals.


\textbf{Hand-drawn sketches:} Sketch-n-Sketch is a bi-directional editing system where direct manipulations to a program's output automatically propagate to the program source code~\citep{Hempel:2016:SSP:2984511.2984575}. This work  compliments our own: programs produced by our method could be provided to a Sketch-n-Sketch-like system as a starting point for further editing.
Other systems in the computer graphics literature convert sketches to procedural representations, using a convolutional network to match a sketch to the output of a parametric 3D modeling system in~\citep{huang2017shape} or supporting interactive sketch-based instantiation of procedural primitives in~\citep{Nishida:2016:ISU:2897824.2925951}. In contrast, we seek to automatically infer a programmatic representation capturing higher-level visual patterns.
The CogSketch system~\citep{forbus2011cogsketch} also aims to have a high-level understanding of hand-drawn figures. Their  goal is cognitive modeling, whereas we are interested in building an automated AI application.

\section{Contributions}

We have presented a system for inferring graphics programs which generate \LaTeX-style figures from hand-drawn images. The system uses a combination of deep neural networks and stochastic search to parse drawings into symbolic specifications; it then feeds these specs to a general-purpose program synthesis engine to infer a structured graphics program. We evaluated our model's performance at parsing novel images, and we demonstrated its ability to extrapolate from provided drawings.
In the near future, we believe it will be possible to produce professional-looking figures just by drawing them and then letting an artificially-intelligent agent write the code.
More generally, we believe the problem of inferring visual programs is a promising direction for research in machine perception.

\subsubsection*{Acknowledgments} We are grateful for advice from Will Grathwohl and Jiajun Wu on  the neural architecture. Funding from NSF GRFP, NSF Award 
\#1753684, the MUSE program (DARPA grant FA8750-14-2-0242),
and AFOSR award FA9550-16-1-0012.

\bibliographystyle{unsrt} 
{\small \bibliography{main}}

\appendix

\section{Appendix}

\subsection{Neural architecture details}\label{architectureDetails}

\subsubsection{Convolutional network}
The convolutional network takes as input 2 $256\times 256$ images
represented as a $2\times 256\times 256$ volume. These are
passed through two layers of convolutions separated by ReLU
nonlinearities and max pooling:
\begin{itemize}
\item Layer 1: 20 $8\times 8$ convolutions, 2 $16\times 4$ convolutions, 2 $4\times 16$ convolutions. Followed by $8\times 8$ pooling with a stride size of 4.
\item Layer 2: 10 $8\times 8$ convolutions. Followed by $4\times 4$ pooling with a stride size of 4.
\end{itemize}

\subsubsection{Autoregressive decoding of drawing commands}

Given the image features $f$, we predict the first token (i.e., the name of the drawing command: \verb|circle|, \verb|rectangle|, \verb|line|, or \verb|STOP|) using logistic regression:
\begin{equation}
  \probability [t_1]\propto \exp\left( W_{t_1}f + b_{t_1}\right)
\end{equation}
where $W_{t_1}$ is a learned weight matrix and  $b_{t_1}$ is a learned bias vector.

Given an attention mechanism $a(\cdot | \cdot)$, subsequent tokens are predicted as:
\begin{equation}
  \probability [t_n|t_{1:(n - 1)}]\propto \text{MLP}_{t_1,n}(a(f|t_{1:(n - 1)}) \oplus \bigoplus_{j < n} \text{oneHot}(t_j))\label{discreteTokenPrediction}
\end{equation}
Thus each token of each drawing primitive has its own learned MLP.
For predicting the coordinates of lines we found that using 32 hidden nodes with sigmoid activations worked well;
for other tokens the MLP's are just logistic regression (no hidden nodes).

We use Spatial Transformer Networks~\cite{jaderberg2015spatial}
as our attention mechanism.
The parameters of the spatial transform are predicted on the basis of previously predicted tokens.
For example, in order to decide where to focus our attention when predicting the $y$ coordinate of a circle,
we condition upon both the identity of the drawing command (\verb|circle|) and upon the value of the previously predicted $x$ coordinate:
\begin{equation}
  a(f|t_{1:(n - 1)}) = \text{AffineTransform}(f, \text{MLP}_{t_1,n}(\bigoplus_{j < n}\text{oneHot}(t_j)))
  \label{spatialTransformEquation}
\end{equation}
So, we learn a different network for predicting special transforms
\emph{for each drawing command} (value of $t_1$) and also \emph{for each token of the drawing command}.
These networks ($\text{MLP}_{t_1,n}$ in equation~\ref{spatialTransformEquation}) have no hidden layers and
output the 6 entries of an affine transformation matrix; see~\cite{jaderberg2015spatial}
for more details.

Training takes a little bit less than a day on a Nvidia TitanX GPU.
The network was trained on $10^5$ synthetic examples.

\subsection{LSTM Baseline for Parsing into Specs}\label{captioningBaseline}
We  compared our deep network with a baseline that models the problem as a kind of image captioning.
Given the target image, this baseline produces the program spec in one shot by
using a CNN to extract features of the input which are passed to an LSTM which finally predicts
the spec token-by-token.
This general architecture is used in several successful neural models of image captioning (e.g.,~\cite{vinyals2015show}).

Concretely, we kept the image feature extractor architecture (a CNN) as in our model,
but only passed it one image as input (the target image to explain).
Then, instead of using an autoregressive decoder to predict a single drawing command,
we used an LSTM to predict a sequence of drawing commands token-by-token.
This LSTM had 128 memory cells,
and at each time step produced as output the next token in the sequence of drawing commands.
It took as input both the image representation and its previously predicted token.

\subsection{Architecture and training of $L_{\text{learned}}$}\label{distanceAppendix}
Our architecture for
$L_{\text{learned}}(\text{render}(S_1)|\text{render}(S_2))$ has the
same series of convolutions as the network that predicts the next
drawing command. We train it to predict two scalars: $|S_1 - S_2|$ and
$|S_2 - S_1|$.  These predictions are made using linear regression
from the image features followed by a ReLU nonlinearity; this
nonlinearity makes sense because the predictions can never be negative
but could be arbitrarily large positive numbers.

We train this network by sampling random synthetic scenes for $S_1$,
and then perturbing them in small ways to produce $S_2$.
We minimize the squared loss between the network's prediction and the ground truth symmetric differences.
$S_1$ is rendered in the ``simulated hand drawing'' style (Section~\ref{generalizingTheHandDrawings}).

\subsection{Simulating hand drawings}\label{noisyAppendix}
We introduce noise into the \LaTeX~rendering process by:
\begin{itemize}
\item Rescaling the image intensity by a factor chosen uniformly at random from $[0.5,1.5]$
\item Translating the image by $\pm 3$ pixels chosen uniformly random
\item Rendering the \LaTeX~using the \verb|pencildraw| style,
  which adds random perturbations to the paths drawn by \LaTeX in a way designed to resemble a pencil.
\item Randomly perturbing the positions and sizes of primitive  \LaTeX drawing commands
\end{itemize}
Empirically this noise process is close enough to the kinds of variations introduced by
an actual hand drawing that
the learned model generalizes to
our test set of hand drawings,
\emph{despite} having never been trained on any real hand drawings.

\subsection{A cost function over programs}\label{costAppendix}
Programs incur a cost of 1 for each command (primitive drawing action,
loop, or reflection).  They incur a cost of $\frac{1}{3}$ for each
unique coefficient they use in a linear transformation beyond the
first coefficient. This encourages reuse of coefficients, which leads
to code that has translational symmetry; rather than provide a
translational symmetry operator as we did with reflection, we modify
what is effectively a prior over the space of program so that it tends
to produce programs that have this symmetry.

Programs also incur a cost of 1 for having loops of constant length 2;
otherwise there is often no pressure from the cost function to explain
a repetition of length 2 as being a reflection rather a loop.

\subsection{Training a search policy}\label{policyAppendix}
Recall from the main paper that our goal is to estimate the policy minimizing the following loss:
\begin{align}
\textsc{Loss}(\theta ; \mathcal{D})& =  \expect_{S\sim\mathcal{D}}\left[ \min_{\sigma\in \text{\textsc{Best}}(S)}\frac{t(\sigma | S)}{\pi_\theta (\sigma | S)}\right] + \lambda \Vert\theta\Vert_2^2\label{policyLoss}\\
\text{where }  \sigma \in\text{\textsc{Best}}(S) &\text{ if  a minimum cost program for }S \text{ is in }\sigma .\nonumber 
\end{align}
We make this optimization problem tractable by annealing our loss function during gradient descent:
\begin{align}
&  \textsc{Loss}_\beta(\theta ; \mathcal{D}) =  \expect_{S\sim\mathcal{D}}\left[ \textsc{SoftMinimum}_\beta \left\{
    \frac{t(\sigma | S)}{\pi_\theta (\sigma | S)}    : \sigma\in \text{\textsc{Best}}(S)\right\}\right] + \lambda \Vert\theta\Vert_2^2\label{softenedObjective}\\
  \text{where }&\textsc{SoftMinimum}_\beta (x_1,x_2,x_3,\cdots ) = \sum_n x_n  \frac{e^{-\beta x_n}}{\sum_{n'}e^{-\beta x_{n'}}}
\end{align}
Notice that $\textsc{SoftMinimum}_{\beta = \infty}(\cdot )$ is just $\min(\cdot )$.
We set the regularization coefficient $\lambda = 0.1$ and minimize equation~\ref{softenedObjective}
using Adam for 2000 steps, linearly increasing $\beta$ from $1$ to $2$.

\subsection{Program synthesis baselines}
\subsubsection{DeepCoder}\label{dcAppendix}
We compared our synthesis policy with a DeepCoder-style baseline.
DeepCoder (DC)~\cite{BalGauBroetal16}
is an approach for learning to speed up program synthesizers.
DC models are neural networks that
predict, starting from a spec,
the probability of a DSL component being in a minimal-cost program satisfying the spec.
Writing $\text{DC}(S)$ for the distribution predicted by the neural network,
DC is trained to maximize  the following objective:
\begin{equation}
  \expect_{S\sim\mathcal{D}}\left[\min_{p\in \text{\textsc{Best}}(S)} \sum_{x\in \text{DSL}} \log \left(\indicator\left[x\in p \right] \text{DC}(S)_x + \indicator\left[x\notin p \right] (1 - \text{DC}(S)_x)\right)
    \right]
\end{equation}
where $x$ ranges over DSL components and $\text{DC}(S)_x\in \left[0,1 \right]$  is the probability predicted by the DC model for component $x$ for spec $S$.

We provided our DC model with the same features given to our bias optimal search policy ($\phi_{spec}$ in Section~\ref{learningASearchPolicy}), used the same log-linear model, and trained using the same 20-fold cross validation splits.
To evaluate the DC baseline on held out data,
we used the \emph{Sort-and-Add} policy described in the DeepCoder paper~\cite{BalGauBroetal16}.

\subsubsection{End--to--End}\label{eeAppendix}

Recall that we factored the graphics program synthesis problem into two components: (1) a perception-facing component, whose job is to go from perceptual input to a set of commands that must occur in the execution of the program (\textbf{spec}); and (2) a program synthesis component, whose job is to infer a program whose execution contains those commands. This is a different approach from other recent program induction models (e.g.,~\cite{devlin2017robustfill,nps}), which regress directly from a program induction problem to the source code of the program.

\textbf{Experiment.} To test whether this factoring is necessary for our domain, we trained a model to regress directly from images to graphics programs. This baseline model, which we call the \emph{no-spec baseline}, was able to infer some simple programs, but failed completely on more sophisticated scenes.

Baseline model architecture: The model architecture is a straightforward, image-captioning-style CNN$\to$LSTM. We keep the same CNN architecture from our main model, with the sole difference that it takes only one image as input. The LSTM decoder produces the program token-by-token: so we flatten the program's hierarchical structure, and use special ``bracketing'' symbols to convey nesting structure, in the spirit of~\citep{vinyals2015grammar}. The LSTM decoder has 2 hidden layers with 1024 units. We used 64-dimensional embeddings for the program tokens.

Training and evaluation: The model was trained on $10^7$ synthetically generated programs -- 2 orders of magnitude more data than the model we present in the main paper. We then evaluated the baseline on \emph{synthetic renders} of our 100 hand drawings (the testing set used throughout the paper). Recall that our model was evaluated on noisy real hand drawings. We sample programs from this baseline model conditioned on a synthetic render of a hand drawing, and report only the sampled program whose output most closely matched the ground truth spec spec, as measured by the symmetric difference of the two sets. We allow the baseline model to spend 1 hour drawing samples per drawing -- recall that our model finds 58\% of programs in under a minute. Together these training and evaluation choices are intended to make the problem as easy as possible for the baseline.

\begin{figure}[h]\centering
      \includegraphics[width = 2cm]{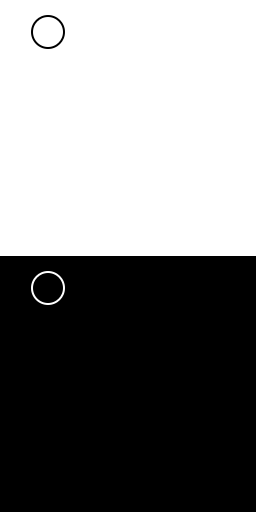}
  \includegraphics[width = 2cm]{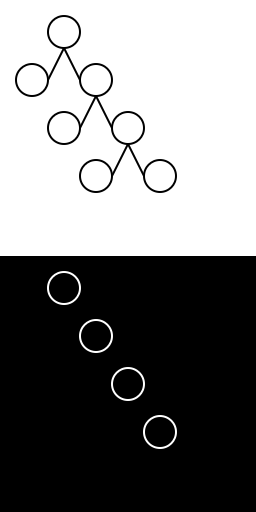}
    \includegraphics[width = 2cm]{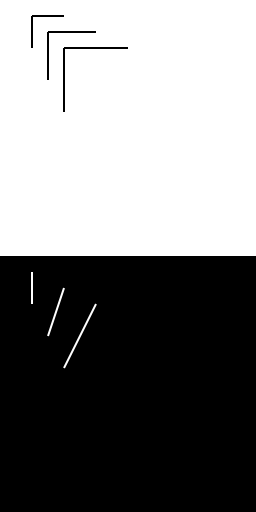}
    \includegraphics[width = 2cm]{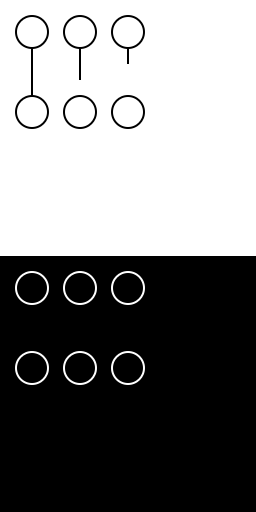}
        \includegraphics[width = 2cm]{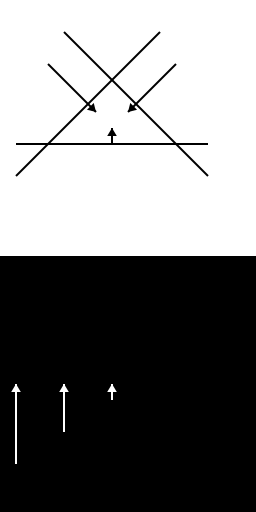}
      \includegraphics[width = 2cm]{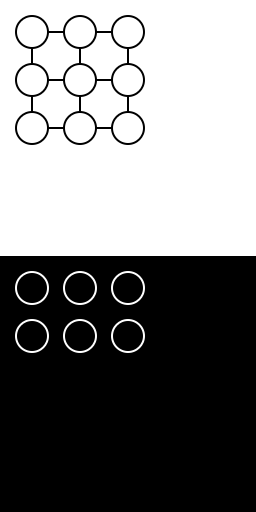}
  \caption{Top, white: synthetic rendering of a hand drawing. Bottom, black: output of best program found by no-spec baseline.}\label{noSpec}
  \end{figure}

Results: The no-spec baseline succeeds for trivial programs (a few lines, no variables, loops, etc.); occasionally gets small amounts of simple looping structure; and fails utterly for most of our test cases. See Figure~\ref{noSpec}.

\end{document}